\documentclass{bmvc2k}

\usepackage{multirow}
\usepackage{graphicx} \graphicspath{ {pictures/} }
\usepackage{caption,subcaption}

\title{Learning to Doodle with Deep Q-Networks and Demonstrated Strokes}

\addauthor{Tao Zhou}{taozhou@cs.ucla.edu}{1}
\addauthor{Chen Fang}{cfang@adobe.com}{2}
\addauthor{Zhaowen Wang}{zhawang@adobe.com}{2}
\addauthor{Jimei Yang}{jimyang@adobe.com}{2}
\addauthor{Byungmoon Kim}{bmkim@adobe.com}{2}
\addauthor{Zhili Chen}{zlchen@adobe.com}{2}
\addauthor{Jonathan Brandt}{jbrandt@adobe.com}{2}
\addauthor{Demetri Terzopoulos}{dt@cs.ucla.edu}{1}

\addinstitution{
 University of California, Los Angeles\\
 Computer Science Department\\
 Los Angeles, CA 90095, USA
}
\addinstitution{
 Adobe Research\\
 345 Park Avenue,\\
 San Jose, USA
}

\runninghead{T.~Zhou}{Learning to Doodle}

\begin{document}

\maketitle

\begin{abstract}
Doodling is a useful and common intelligent skill that people can learn and master. In this work, we propose a two-stage learning framework to teach a machine to doodle in a simulated painting environment via Stroke Demonstration and deep Q-learning (SDQ). The developed system, \emph{Doodle-SDQ}, generates a sequence of pen actions to reproduce a reference drawing and mimics the behavior of human painters. In the first stage, it learns to draw simple strokes by imitating in supervised fashion from a set of stroke-action pairs collected from artist paintings. In the second stage, it is challenged to draw real and more complex doodles without ground truth actions; thus, it is trained with Q-learning. Our experiments confirm that (1) doodling can be learned without direct step-by-step action supervision and (2) pretraining with stroke demonstration via supervised learning is important to improve performance. We further show that \emph{Doodle-SDQ} is effective at producing plausible drawings in different media types, including sketch and watercolor. A short video can be found at \url{https://www.youtube.com/watch?v=-5FVUQFQTaE}.
\end{abstract}

\section{Introduction}
\label{sec:intro}

Doodling is a common, simple, and useful activity for communication, education, and reasoning. It is sometimes very effective at capturing complex concepts and conveying complicated ideas~\cite{brown2014doodle}. Doodling is also quite popular as a simple form of creative art, compared to other types of fine art. We all learn, practice, and master the skill of doodling in one way or another. Therefore, for the purposes of building a computer-based doodling tool or enabling computers to create art, it is interesting and meaningful to study the problem of teaching a machine to doodle.

\begin{figure}
\centering
\includegraphics[width=0.225\linewidth]{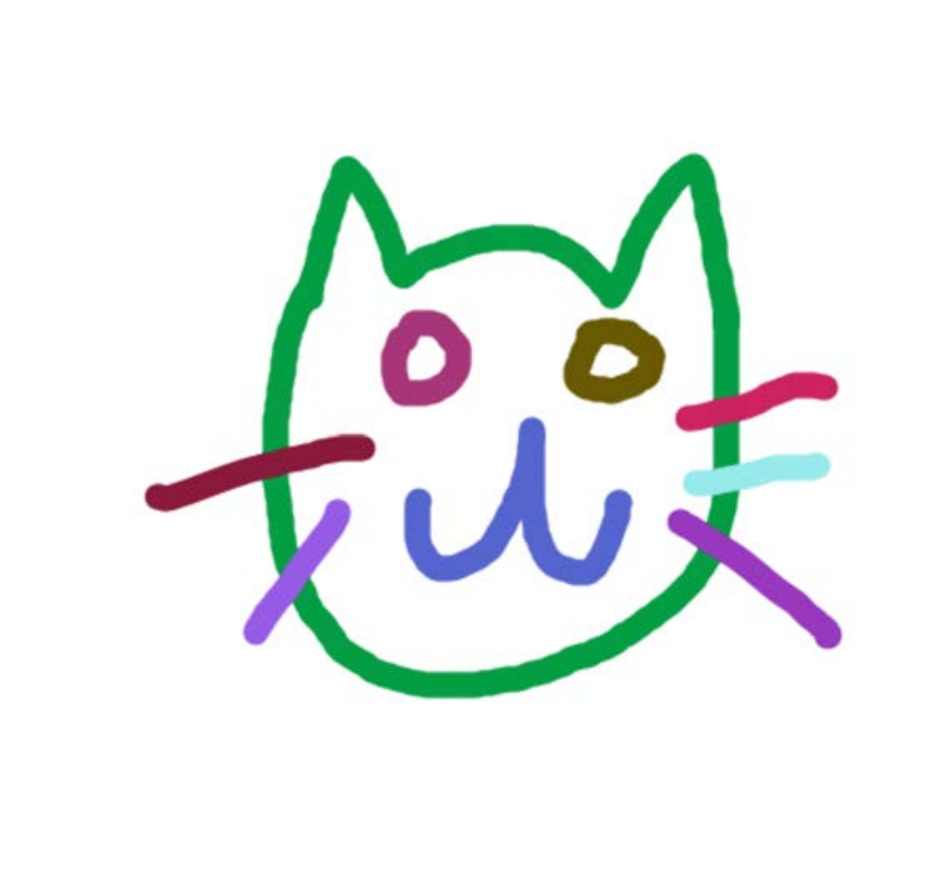}
\includegraphics[width=0.225\linewidth]{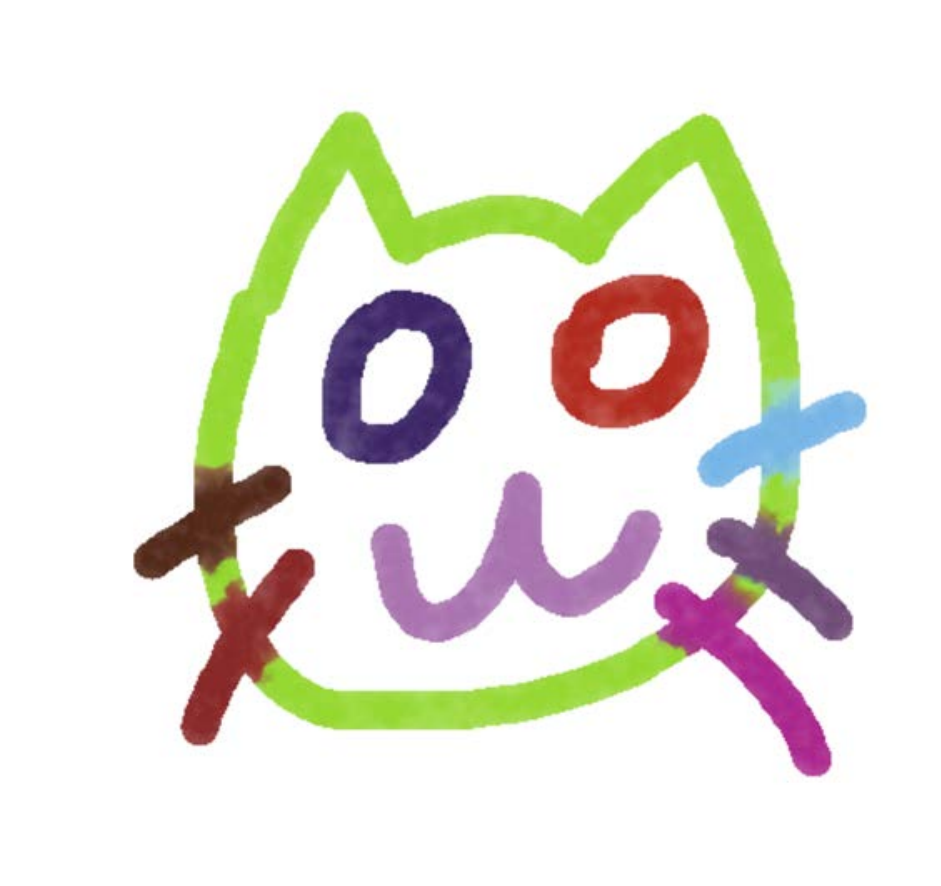}
\caption{Cat doodles rendered using color sketch (left) and water color (right) media types.}
\label{fig:mediatype}
\end{figure}

Recent progress in visual generative models---e.g., Generative Adversarial Networks~\cite{goodfellow2014generative} and Variational Autoencoders~\cite{kingma2013auto}---have enabled computer programs to synthesize complex visual patterns, such as natural images~\cite{denton2015deep}, videos~\cite{vondrick2016generating}, and visual arts~\cite{elgammal2017can}. By contrast to these efforts, which model pixel values, we model the relationship between pen actions and visual outcomes, and use that to generate doodles by acting in a painting environment. More concretely, given a reference doodle drawing, our task is to doodle in a painting environment so as to generate a drawing that resembles the reference. In order to facilitate the experiment setup and be more focused and expedient on algorithm design, we employ an internal Simulated Painting Environment (SPE) that supports major media types; for example, sketch and watercolor (Figure~\ref{fig:mediatype}).

Our seemingly simple task faces at least \textit{three challenges}:

First, our goal is to enable machines to doodle like humans. This means that rather than mechanically printing pixel by pixel like a printer, our system should be able to decompose a given drawing into strokes, assign them a drawing order, and reproduce the strokes with pen action sequences. These abilities require the system to visually parse the given drawing, understand the current status of the canvas, make and adjust drawing plans, and implement the plans by invoking correct actions in a painting environment. Rather than designing a rule-based or heuristic system that is likely to fail in corner cases, we propose a machine learning framework for teaching computers to accomplish these tasks.

The second challenge is the lack of data to train such a system. The success of modern machine learning heavily relies on the availability of large-scale labeled datasets. However, in our domain, it is expensive, if not impossible, to collect paintings and their corresponding action data (i.e., recordings of artists' actions). This is compounded by the fact that the artistic paintings space features rich variations, including media types, brush settings, personal styles, etc., that are difficult to cover. Hence, the traditional paradigm of collecting ground truth data for model learning does not work in our case. 

Consequently, we propose a hybrid learning framework that consists of two stages of training, which are driven by different learning mechanisms. In Stage~1, we collect stroke demonstration data, which comprises a picture of randomly placed strokes and its corresponding pen actions recorded from a painting device, and train a model to draw simple strokes in a supervised manner. Essentially, the model is trained to imitate human drawing behaviour at the stroke level with step-by-step supervision. Note that it is significantly easier to collect human action data at the stroke level than for the entire painting. In Stage~2, we challenge the model learned in Stage~1 with real and more complex doodles, for which there are no associated pen action data. To train the model, we adopt a Reinforcement Learning (RL) paradigm, more specifically Q-learning with reward for reproducing a given reference drawing. We name our proposed system \emph{Doodle-SDQ}, which stands for Doodle with Stroke Demonstration and deep Q-Networks. We experimentally show that both stages are required to achieve good performance.

Third, it is challenging to induce good painting behaviour with RL due to the large state/action space. At each step, the agent faces at least 200 different action choices, including the pen state, pen location, and color. The action space is larger than in other settings where RL has been applied successfully~\cite{mnih2015human,peng2016terrain,levine2016end}. We empirically observe that Q-learning with a high probability of random exploration is not effective in our large action space, and reducing the chance of random exploration significantly helps stabilize the training process, thus improving the accumulated reward.

\begin{figure}
\begin{center}
\includegraphics[width=0.1\linewidth,trim={0 0 6cm 0},clip]{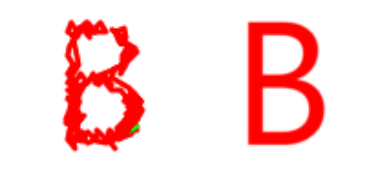}
\includegraphics[width=0.1\linewidth,trim={0 0 6cm 0},clip]{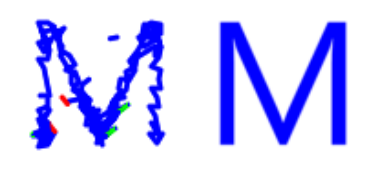} 
\includegraphics[width=0.1\linewidth,trim={0 0 6cm 0},clip]{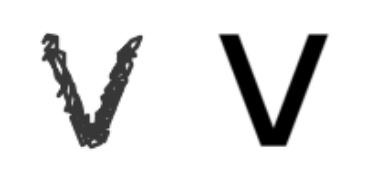} 
\includegraphics[width=0.1\linewidth,trim={0 0 6cm 0},clip]{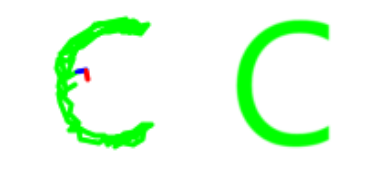}\\
\includegraphics[width=0.1\linewidth,trim={6cm 0 0 0},clip]{colors/B2}
\includegraphics[width=0.1\linewidth,trim={6cm 0 0 0},clip]{colors/13} 
\includegraphics[width=0.1\linewidth,trim={6cm 0 0 0},clip]{binary/jerk} 
\includegraphics[width=0.1\linewidth,trim={6cm 0 0 0},clip]{colors/0}
\end{center}
\caption{Sketch Drawing Examples: \textbf{BMVC}. (top) The images produced by unrolling the Doodle-SDQ model for 100 steps. (bottom) The corresponding reference images.}
\label{fig:BMVC}
\end{figure}

To summarize, Doodle-SDQ leverages demonstration data at the stroke level and generates a sequence of pen actions given only reference images. Our algorithm models the relationship between pen actions and visual outcomes and works in a relatively large action space. We apply our trained model to draw various concepts (e.g., characters and objects) in different media types (e.g., black and white sketch, color sketch, and watercolor). In Figure~\ref{fig:BMVC}, our system has automatically sketched a colored ``BMVC''.

\section{Related Work}

\subsection{Imitation Learning and Deep Reinforcement Learning} 

Imitation learning techniques aim to mimic human behavior in a given task. An agent (a learning machine) is trained to perform a task from demonstrations by learning a mapping between observations and actions~\cite{hussein2017imitation}.
Naive imitation learning, however, is unable to help the agent recover from its mistakes, and the demonstrations usually cannot cover all the scenarios the agent will experience in the real world. To tackle this problem, DAGGER~\cite{ross2011reduction} iteratively produces new policies based on polling the expert policy outside its original state space. Therefore, DAGGER requires an expert to be available during training to provide additional feedback to the agent. When the demonstration data or the expert are unavailable, RL is a natural choice for an agent to learn from experience by exploring the world. Nevertheless, reward functions have to be designed based on a large number of hand-crafted features or rules~\cite{xie2012artist}.

The breakthrough of Deep RL (DRL)~\cite{mnih2015human} came from the introduction of a target network to stabilize the training process and experience replay to learn from past experiences. \citet{van2016deep} proposed \textit{Double DQN} (DDQN) to solve an over-estimation issue in deep Q-learning due to the use of the maximum action value as an approximation to the maximum expected action value. \citet{schaul2016prioritized} developed the concept of \textit{prioritized experience replay}, which replaced DQN's uniform sampling strategy from the replay memory with a sampling strategy weighted by TD errors. Our algorithm starts with Double DQN with prioritized experience replay (DDQN $+$ PER) \cite{schaul2016prioritized}.

Recently, there has also been interest in combining imitation learning with the RL problem~\cite{cruz2017pre,subramanian2016exploration}. \citet{silver2016mastering} trained human demonstrations in supervised learning and used the supervised learner's network to initialize RL's policy network while \citet{hester2018deep} proposed Deep Q-learning from Demonstrations (DQfD), which leverages even very small amounts of demonstration data to accelerate learning dramatically.

\subsection{Sketch and Art Generation} 

There are outstanding studies related to drawing in the fields of robotics and AI. Traditionally, a robot arm is programmed to sketch lines on a canvas to mimic a given digitized portrait~\cite{tresset2013portrait}. Calligraphy skills can be acquired via Learning from Demonstration~\cite{sun2014robot}. Recently, Deep Neural Network-based approaches for art generation have been developed~\cite{gatys2016image,elgammal2017can}. An earlier work by \citet{gregor2015draw} introduced a network combining an attention mechanism with a sequential auto-encoding framework that enables the iterative construction of complex images. The high-level idea is similar to ours; that is, updating only part of the canvas at each step. Their method, however, operates on the canvas matrix while ours generates pen actions that make changes to the canvas. More recently, a SPIRAL model~\cite{ganin2018synthesizing} used Reinforced Adversarial Learning to produce impressive drawings without supervision; however, the model generates control points for quadratic Bezier curves, rather than directly controlling the pen's drawing actions.

Rather than focusing on traditional pixel image modeling approaches, \citet{zhang2017drawing} and \citet{simhon2004sketch} proposed generative models for vector images. \citet{graves2013generating} focused on handwriting generation with Recurrent Neural Networks to generate continuous data points. Following the handwriting generation work, a sketch-RNN model was proposed to generate sketches~\cite{ha2017neural,Jongejan2016quick}, which was learned in a fully supervised manner. The features learned by the model were represented as a sequence of pen stroke positions. In our work, we process the sketch sequence data and, using an internal simulated painting environment, render onto the canvas as in the reference images.

\section{Methodology}

Given a reference image and a blank canvas for the first iteration, our Doodle-SDQ model predicts the pen's action. When the pen moves to the next location, a new canvas state is produced. The model takes the new canvas state as the input, predicts the action based on the difference between the current canvas and the reference image, and repeats the process for a fixed number of steps. (Figure~\ref{fig:framework}a).

\subsection{Our Model}

\begin{figure}
\begin{center}
\subcaptionbox{}{\includegraphics[width=0.48\textwidth]{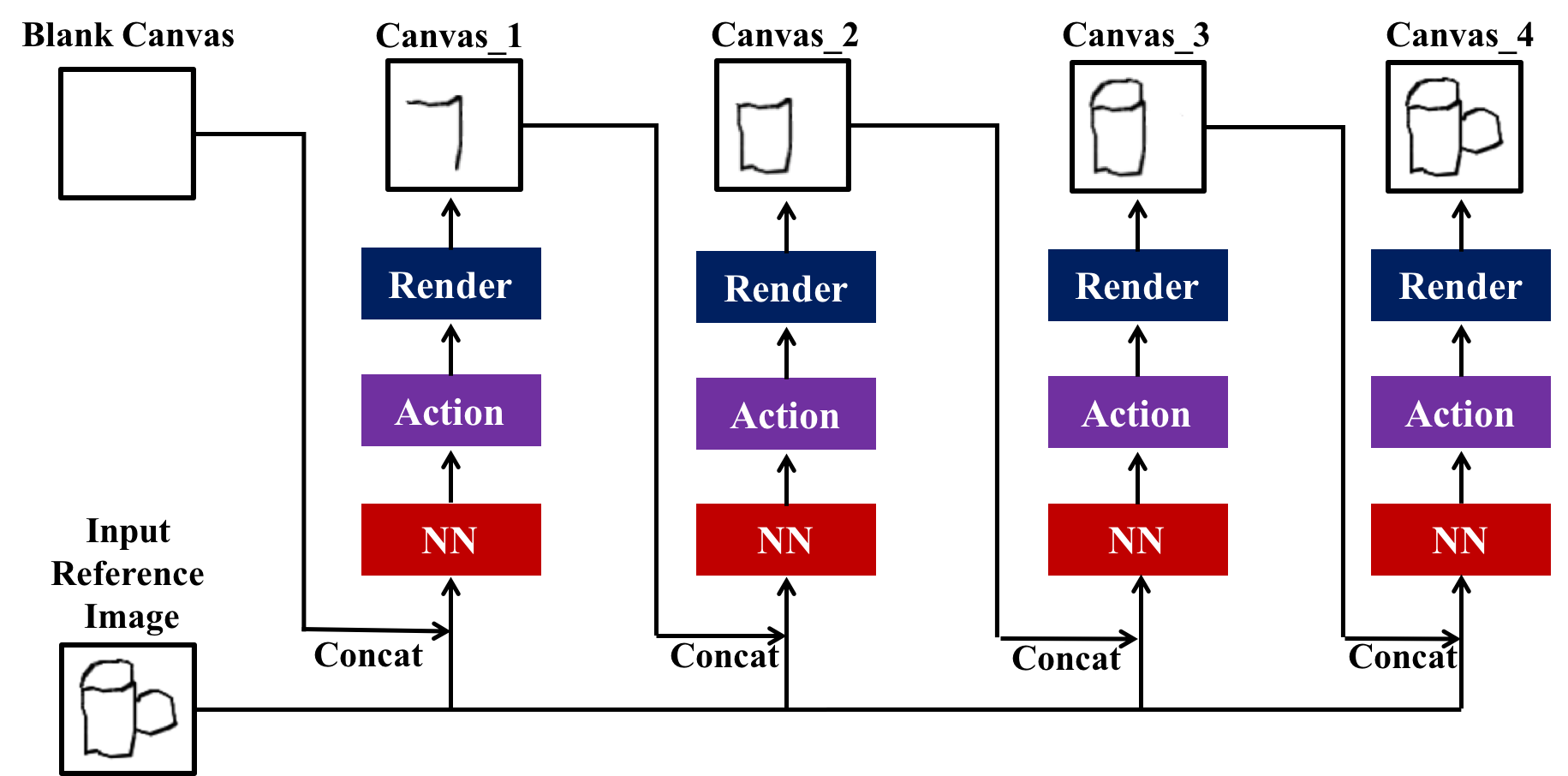}}
\hfill
\subcaptionbox{}{\includegraphics[width=0.48\textwidth]{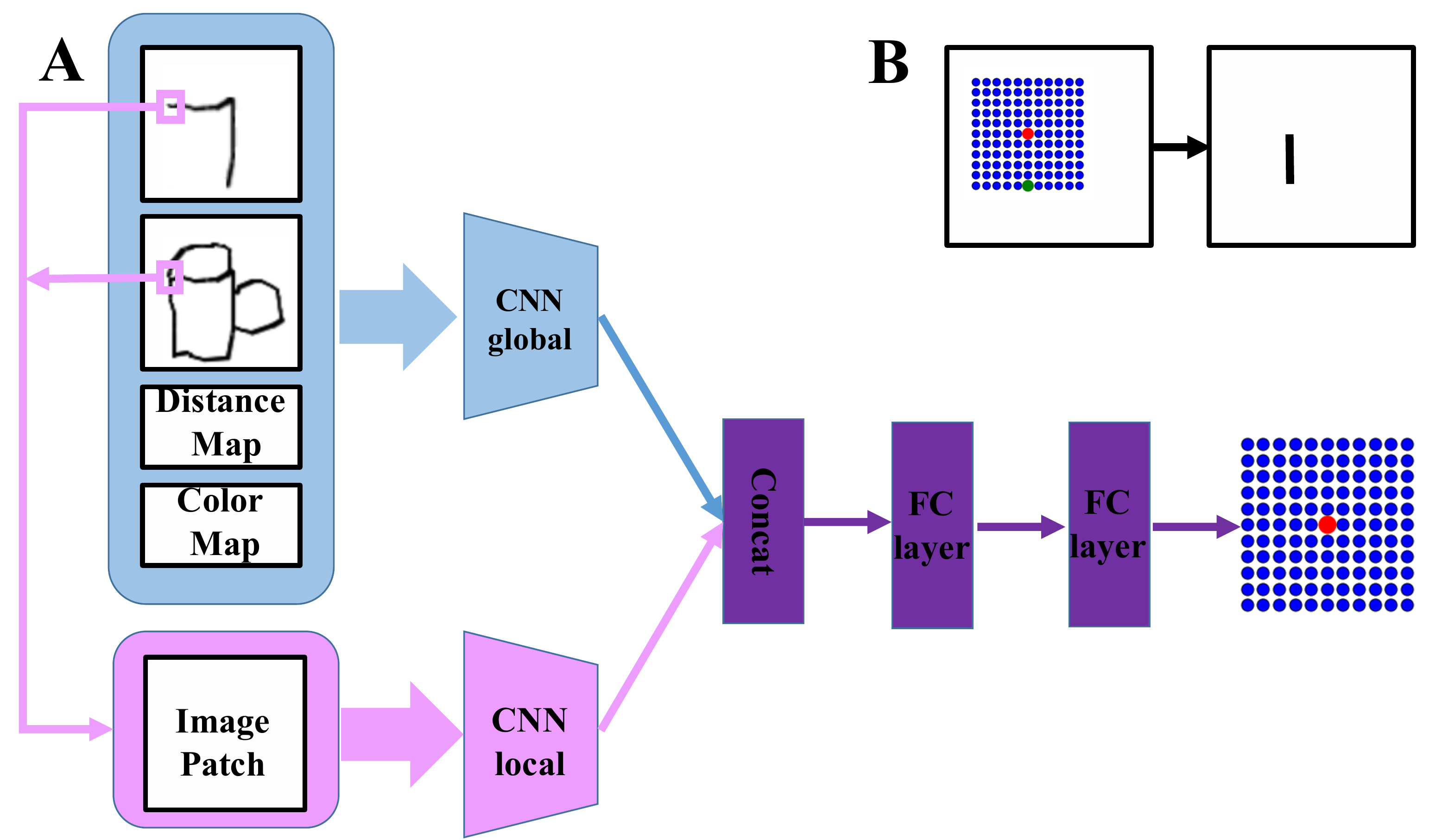}}
\end{center}
\caption{Doodle-SDQ structure. (a) The algorithm starts with a blank canvas and an input reference image. The neural network predicts the action of the pen and sends rendering commands to a painting engine. The new canvas and the reference image are then concatenated and the process is repeated for a fixed number of steps. (b) A: Two CNNs extract global scene-level contextual features and local image patch descriptors. The local and global features are concatenated for action prediction. B: Given the current position (red dot) and the predicted action (green dot), the painting engine renders a segment to connect them. The rectangle of blue dots represents the movement range, which is the same size as the local image patch.}
\label{fig:framework}
\end{figure}

The network has two input streams (Figure~\ref{fig:framework}b-A). The global stream has 4 channels, which comprise the current canvas, the reference image, the distance map and the color map. The distance map and the color map encode the pen's position and state. The local stream has 2 channels---the cropped patch of the current canvas centered at the pen's current location with size equal to the pen's movement range, and the corresponding patch on the reference image. Unlike the classical DQN structure~\cite{mnih2015human}, which stacks four frames, the input in this model includes only the current frame and no history information.

The convnet for global feature extraction consists of three convolutional layers~\cite{mnih2015human}. The first hidden layer convolves 32 $8\times8$ filters with stride 4. The second hidden layer convolves 64 $4\times4$ filters with stride 2. The third hidden layer convolves 64 $3\times3$ filters with stride 1. The only convnet layer of the local CNN stream convolves 128 $11\times11$ filters with stride 1. The two streams are then concatenated and input to a fully-connected linear layer, and the output layer is another fully-connected linear layer with a single output for each valid action. 

At each time step, the pen must decide where to move (Figure~\ref{fig:framework}b-B). The pen is designed to have maximum 5 pixels offset movement horizontally and vertically from its current position.\footnote{The maximal offset movement of the pen is set arbitrarily; it could also be 4 or 6.} Therefore, the movement range is $11\times11$ and there are in total 121 positional choices.

The pen's state is determined by the type of reference image. Specifically, the pen's state is either up or down (i.e., draw) for a grayscale image. For a color image, the pen's state can be up or down with a color selected from the three options (i.e., red, green, and blue).\footnote{The painting engine allows more colors; however, to simplify our experiments, we limit it to three colors.} Therefore, the dimension of the action space is 242 for grayscale images and 484 for color images. Figure~\ref{fig:framework}b-B shows a segment rendered given the pen's current position and the predicted action.

Rather than memorizing absolute coordinates of a pen on a canvas, humans tend to encode the relative positions between points. To represent the current location of the pen, an L2 distance map is constructed by computing
\begin{equation}
D(x, y) = \frac{\sqrt{(x - x_o)^2 + (y - y_o)^2}}{L}, \quad \forall(x,y)\in\Omega,
\end{equation}
where $\Omega$ denotes the canvas which is an $L\times L$ discrete grid, $L$ being the length of the canvas' side, and $(x_o, y_o)$ is the current pen location. In terms of a color map, all elements are 1 when the pen is put down and 0 when the pen is lifted up for grayscale images. For an image with red, green, and blue color, all elements are 0 when the pen is lifted up, 1 for red color drawing, 2 for green color and 3 for blue color. The size of distance map and the color map is the same as the canvas size, which is $84 \times 84$ (Figure~\ref{fig:framework}b-A). Table~\ref{table:input_output} summarizes the dimensionalities of the input and output for grayscale or color reference images.

\begin{table}[ht]
\small
\begin{center}
\begin{tabular}{|l|c|c|c|}
\hline
Image & Input global stream & Input local stream & Output action space\\
\hline
Grayscale & $84\times84\times4$ & $11\times11\times2$ & $11\times11\times2 = 242$\\
RGB & $84\times84\times8$ & $11\times11\times6$ & $11\times11\times4 = 484$\\
\hline
\end{tabular}
\end{center}
\caption{Input and output dimensionalities.}
\label{table:input_output}
\end{table}

\subsection{Pre-Training Networks Using Demonstration Strokes}

\begin{figure}
\begin{center}
\subcaptionbox{}{\includegraphics[width=0.19\textwidth]{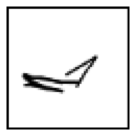}}
\subcaptionbox{}{\includegraphics[width=0.19\textwidth]{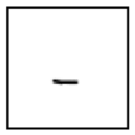}}
\subcaptionbox{}{\includegraphics[width=0.19\textwidth]{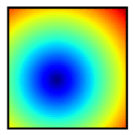}}
\subcaptionbox{}{\includegraphics[width=0.19\textwidth]{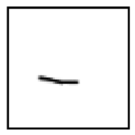}}
\subcaptionbox{}{\includegraphics[width=0.19\textwidth]{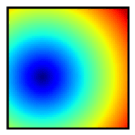}}
\end{center}
\caption{Data preparation for pre-training the network. (a) A reference image comprising two strokes randomly placed on the canvas; (b) the current canvas as part of the reference image; (c) the distance map of the current canvas, whose center is the pen's location on the current canvas; (d) the next step canvas after a one step action of the pen; (e) The distance map of the next step canvas, which represents the pen's location on the next step canvas.}
\label{fig:DP}
\end{figure}

DRL can be difficult to train from scratch. Therefore, we pre-train the network in a supervised manner using synthesized data with ground truth actions. The synthetic data are generated by randomly placing real strokes on canvas (Figure~\ref{fig:DP}a). The real strokes are collected from recordings of a few artist paintings.

In the learning from demonstration phase, each training sample consists of the reference image (Figure~\ref{fig:DP}a), the current canvas (Figure~\ref{fig:DP}b), the color map, the distance map (Figure~\ref{fig:DP}c), the small patch of the reference image, and the current canvas. The ground truth output will be the drawing action producing Figure~\ref{fig:DP}d from Figure~\ref{fig:DP}b. After training, the learned weights and biases are used to initialize the Doodle-SDQ network in the RL stage.

\subsection{Doodle-SDQ}

To encourage the agent to draw a picture similar to the reference image, the similarity between the $k^{\mathrm{\scriptsize th}}$ step canvas and the reference image is measured as
\begin{equation}
s_k = \frac{\mathop{\sum_{i=1}^{L}\sum_{j=1}^{L}}(P_{ij}^{k} -  P_{ij}^{\mathrm{\scriptsize ref}})^2}{L^2},
\end{equation}
where $P_{ij}^{k}$ is the pixel value at position $(i, j)$ in the $k^{\mathrm{\scriptsize th}}$ step canvas and $P_{ij}^{\mathrm{\scriptsize ref}}$ is the pixel value at that position in the reference image.

The pixel reward of executing action at the $k^{\mathrm{\scriptsize th}}$ step is defined as
\begin{equation}
r_{\mathrm{\scriptsize pixel}} = s_k - s_{k+1}.
\end{equation} 
An intuitive interpretation is that $r_{\mathrm{\scriptsize pixel}}$ is 0 when the pen is up and increases with the similarity between the canvas and reference image.

To avoid slow movement or pixel by pixel printing, we penalize small steps. Specifically, if the pen moves less than 5 pixels/step when the pen is drawing or if it moves while being up, the agent will be penalized with $P_{\mathrm{\scriptsize step}}$. If the input is an RGB image, we additionally penalize the incorrectness of the chosen color $P_{\mathrm{\scriptsize color}}$.

Thus, the final reward is 
\begin{equation}
r_k = r_{\mathrm{\scriptsize pixel}} + P_{\mathrm{\scriptsize step}} + \beta P_{\mathrm{\scriptsize color}},
\end{equation} 
where $P_{\mathrm{\scriptsize step}}$ and $P_{\mathrm{\scriptsize color}}$ are constants set based on the observation, and $\beta$ depends on the input image type: 0 for a grayscale image and 1 for a color image. 

\begin{figure}
\includegraphics[width=0.12\linewidth]{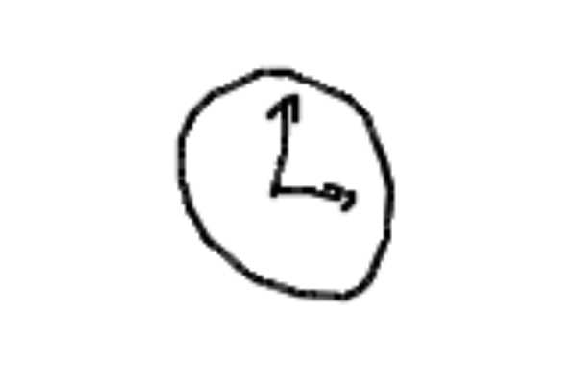}
\includegraphics[width=0.12\linewidth]{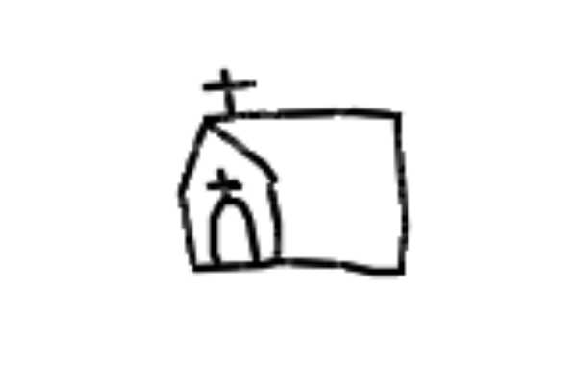}
\includegraphics[width=0.12\linewidth]{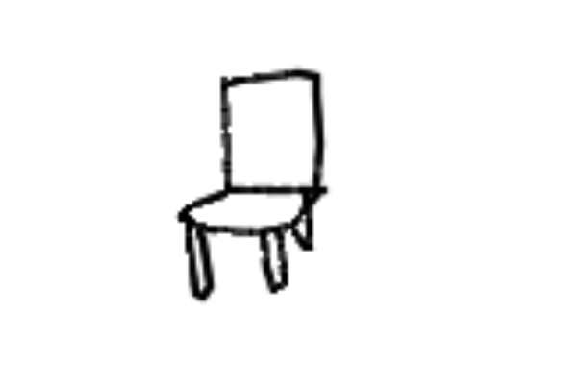}
\includegraphics[width=0.12\linewidth]{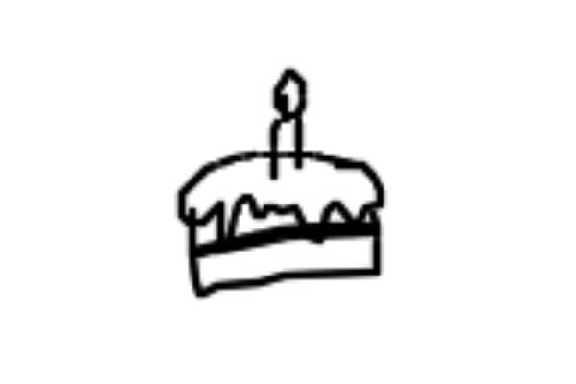}
\includegraphics[width=0.12\linewidth]{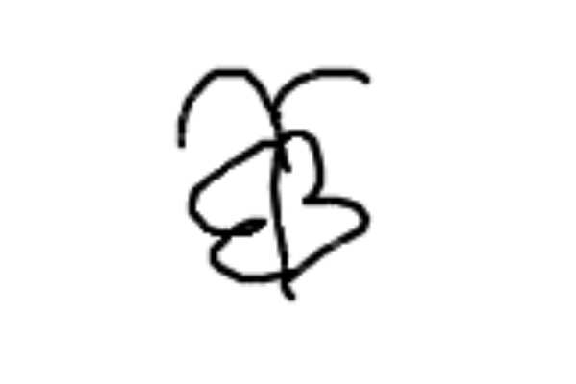}
\includegraphics[width=0.12\linewidth]{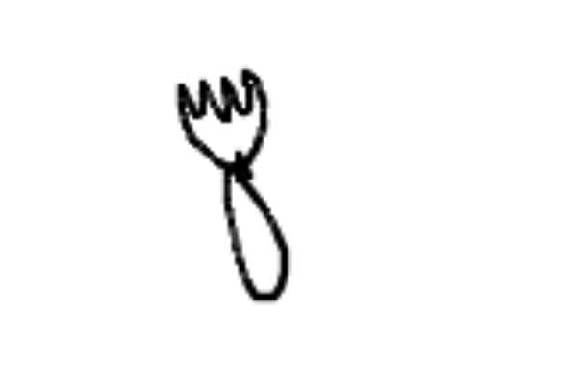} 
\includegraphics[width=0.12\linewidth]{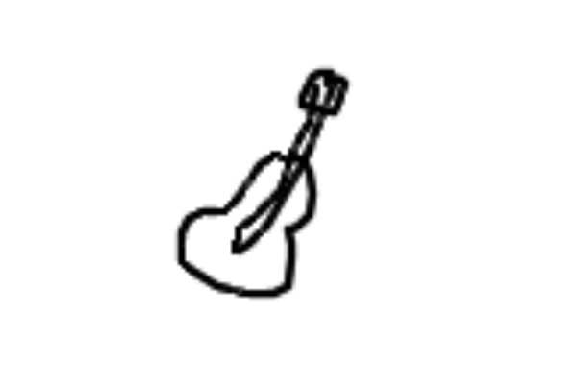}
\includegraphics[width=0.12\linewidth]{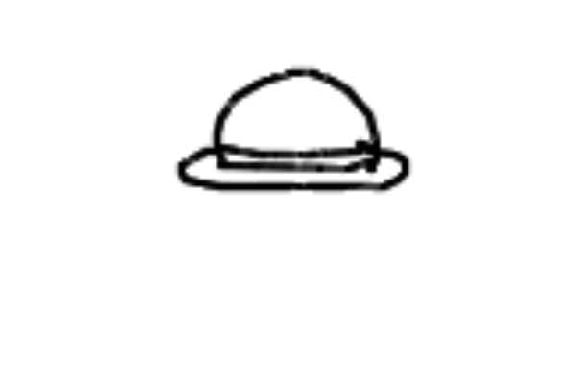}
\includegraphics[width=0.12\linewidth]{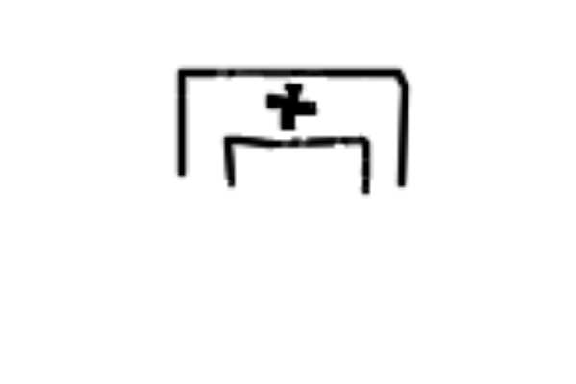}
\includegraphics[width=0.12\linewidth]{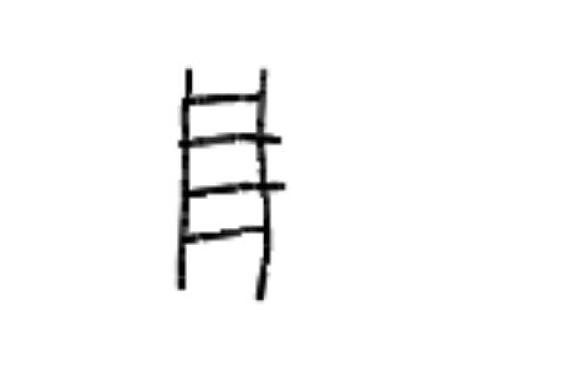}
\includegraphics[width=0.12\linewidth]{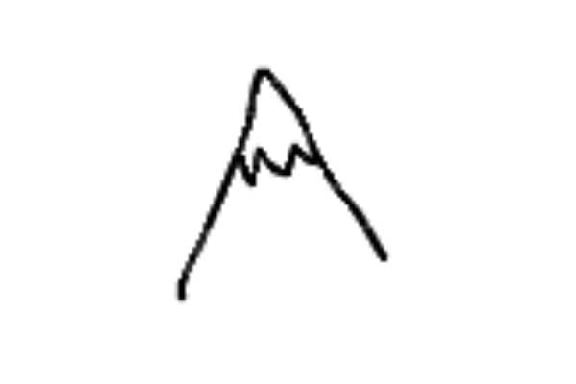}
\includegraphics[width=0.12\linewidth]{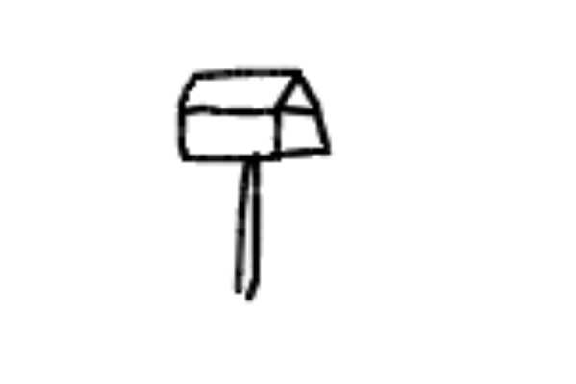}
\includegraphics[width=0.12\linewidth]{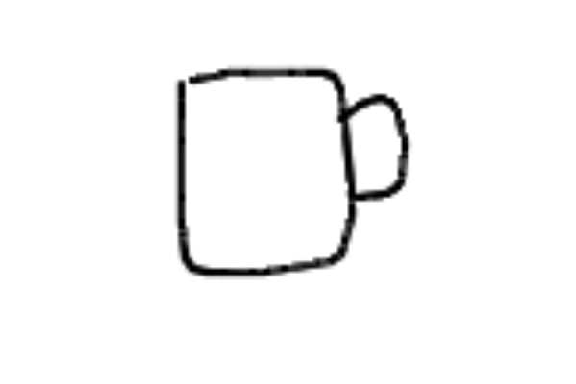}
\includegraphics[width=0.12\linewidth]{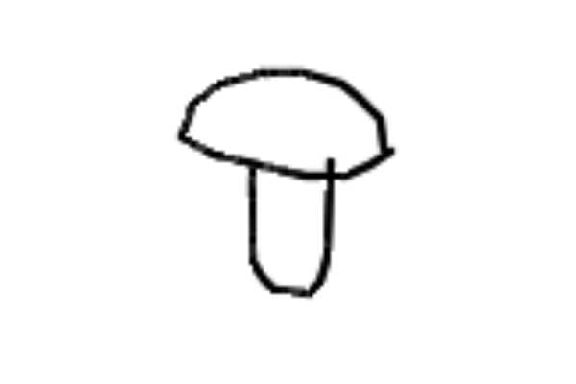}
\includegraphics[width=0.12\linewidth]{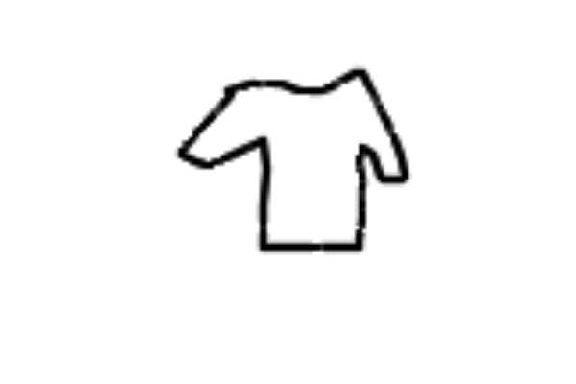}
\includegraphics[width=0.12\linewidth]{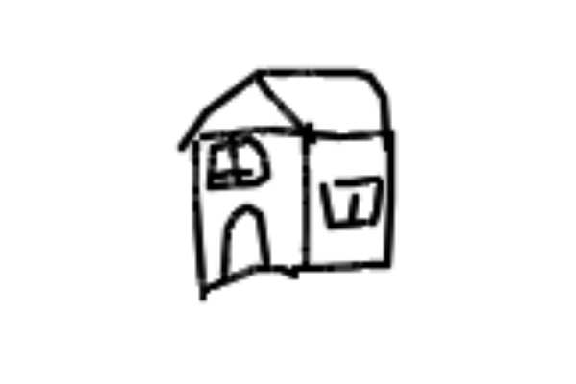}
\caption{Reference images for training and testing. 16 classes are randomly chosen from the QuickDraw dataset~\cite{Jongejan2016quick}: clock, church, chair, cake, butterfly, fork, guitar, hat, hospital, ladder, mountain, mailbox, mug, mushroom, T-shirt, house.}
\label{fig:googledraw}
\end{figure}

In the RL phase, we use QuickDraw~\cite{Jongejan2016quick}, a dataset of vector drawings, as the input reference image. Since the scale of the drawings in QuickDraw varies across samples, the drawing sequence data is processed such that all the drawings can be squeezed onto an $84\times84$ pixel canvas. We randomly selected sixteen classes, and each class includes 200 reference images (Figure~\ref{fig:googledraw}).
For RL training, the images except for the `house' class are applied. Therefore, 3,000 reference images are adopted for training.

\section{Experiments}

During the pretraining phase, we use a softmax cross entropy loss for the classification task. The loss is minimized using Adam~\cite{kingma2014adam} with minibatches of size 128 for optimization with the initial step size $\alpha = 0.001$, and gradually decays with the training step. Instead of using random initialization, the learned weights from the pretrained classification model are used to initialize Doodle-SDQ's network. Due to the large action space, the pen is likely to draw a wrong stroke following a random action in the RL phase. Thus, exploration in action space is rarely applied unless the pen is stuck at some point.\footnote{From our observations, the pen is likely to stop moving at some location or move back and forth between two spots. Only in these scenarios, the pen will be given a random action to avoid local minima.} For the RL stage, we train for a total of 0.6M frames and use a replay memory of 20 thousand frames. The weights are updated based on the difference between the Q value and the output of the target Q network~\cite{schaul2016prioritized}. The loss is minimized using Adam with $\alpha = 0.001$. Our model is implemented in Tensorflow~\cite{abadi2016tensorflow}. We plan to release our code, data, and the painting engine to facilitate the reproduction of our results.

\begin{figure}
\begin{center}
\subcaptionbox{Sketch: butterfly, guitar, church, cake, mailbox, hospital}{
\includegraphics[width=0.16\linewidth]{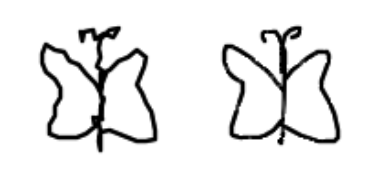} 
\includegraphics[width=0.16\linewidth]{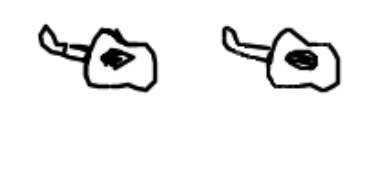} 
\includegraphics[width=0.16\linewidth]{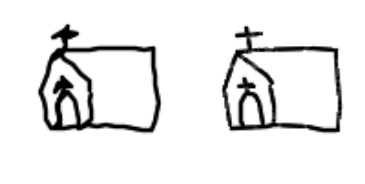}
\includegraphics[width=0.16\linewidth]{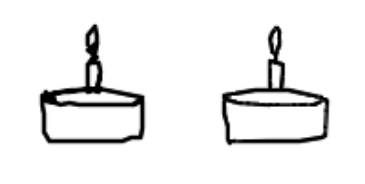}
\includegraphics[width=0.16\linewidth]{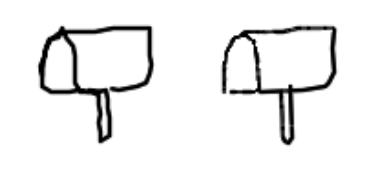}
\includegraphics[width=0.16\linewidth]{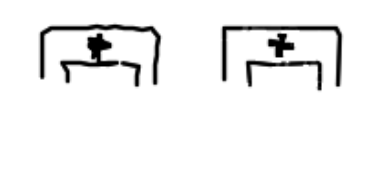}}
\\[5pt]
\subcaptionbox{Color sketch: mailbox, chair, hat, house, mug, T-shirt}{
\includegraphics[width=0.16\linewidth]{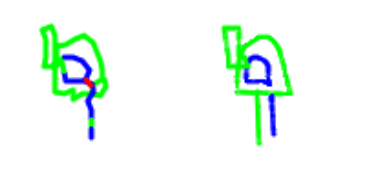} 
\includegraphics[width=0.16\linewidth]{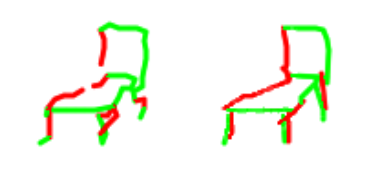} 
\includegraphics[width=0.16\linewidth]{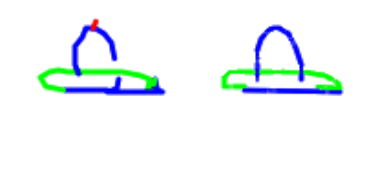}
\includegraphics[width=0.16\linewidth]{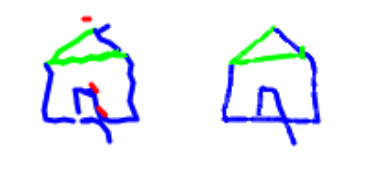}
\includegraphics[width=0.16\linewidth]{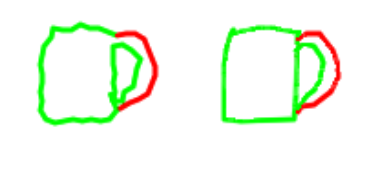}
\includegraphics[width=0.16\linewidth]{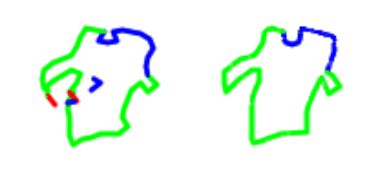}}
\\[5pt]
\subcaptionbox{Watercolor: T-shirt, butterfly, cake, mug, house, mailbox}{
\includegraphics[width=0.325\linewidth]{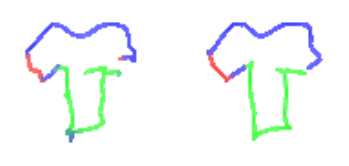} 
\includegraphics[width=0.325\linewidth]{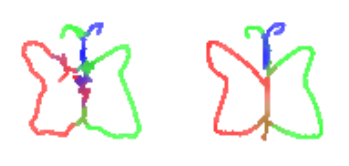}
\includegraphics[width=0.325\linewidth]{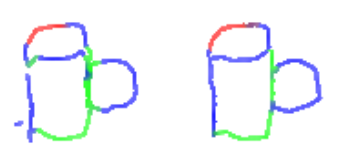}}
\\[5pt]
\end{center}
\caption{Comparisons between drawings and reference images in different media types: (a) sketch, (b) color sketch, (c) watercolor. The left image in each pair is the drawing after 100 steps of the model and the right is the reference image. The drawings in watercolor mode are enlarged to visualize the stroke distortion and color mixing}
\label{fig:result}
\end{figure}

\begin{figure}
\begin{center}
\includegraphics[width=0.16\linewidth]{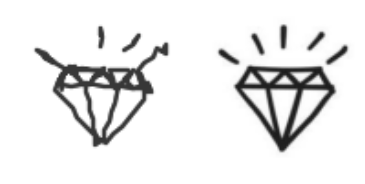} 
\includegraphics[width=0.16\linewidth]{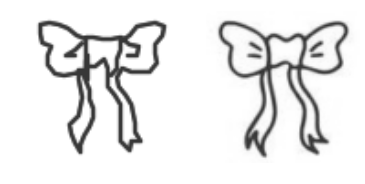} 
\includegraphics[width=0.16\linewidth]{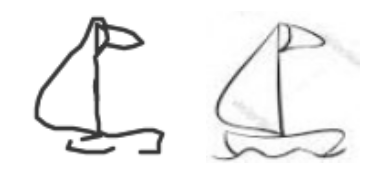}
\includegraphics[width=0.16\linewidth]{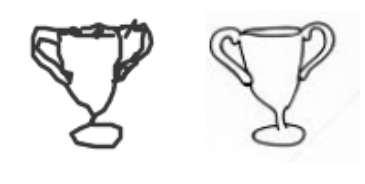} 
\includegraphics[width=0.16\linewidth]{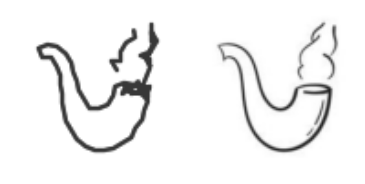} 
\includegraphics[width=0.16\linewidth]{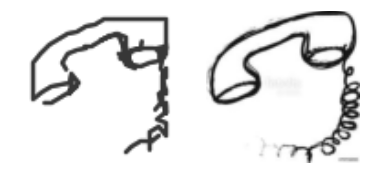}

\includegraphics[width=0.16\linewidth]{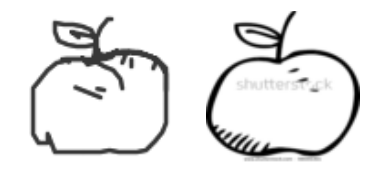} 
\includegraphics[width=0.16\linewidth]{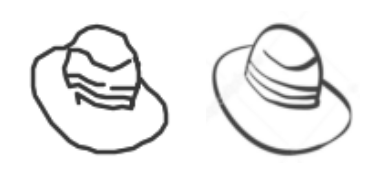} 
\includegraphics[width=0.16\linewidth]{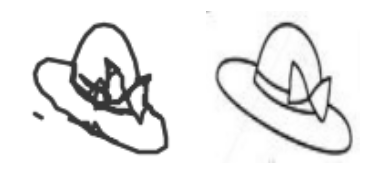}
\includegraphics[width=0.16\linewidth]{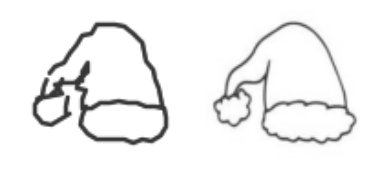} 
\includegraphics[width=0.16\linewidth]{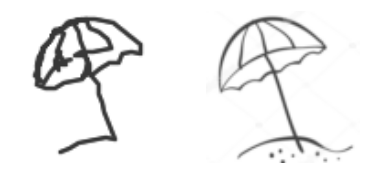}
\includegraphics[width=0.16\linewidth]{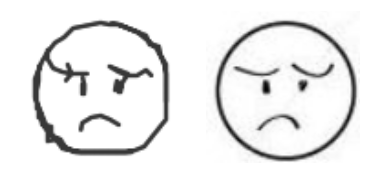} 

\includegraphics[width=0.16\linewidth]{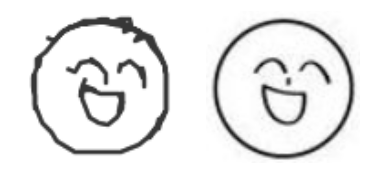} 
\includegraphics[width=0.16\linewidth]{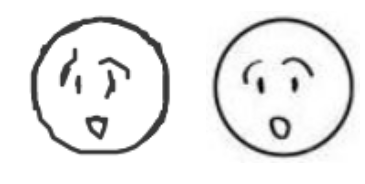}
\includegraphics[width=0.16\linewidth]{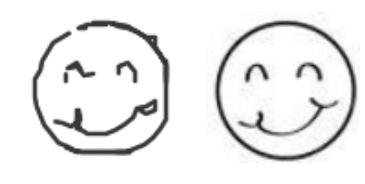}
\includegraphics[width=0.16\linewidth]{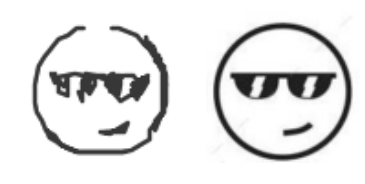} 
\includegraphics[width=0.16\linewidth]{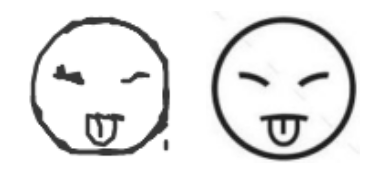} 
\includegraphics[width=0.16\linewidth]{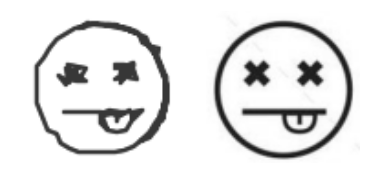}

\includegraphics[width=0.16\linewidth]{colors/B2}
\includegraphics[width=0.16\linewidth]{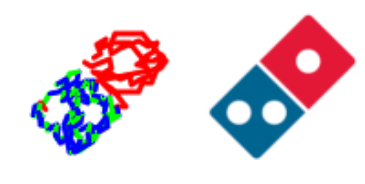}
\includegraphics[width=0.16\linewidth]{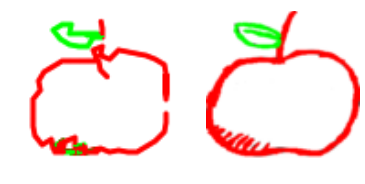} 
\includegraphics[width=0.16\linewidth]{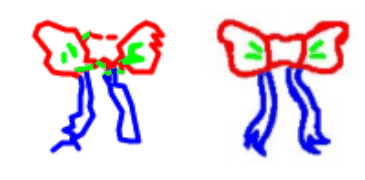} 
\includegraphics[width=0.16\linewidth]{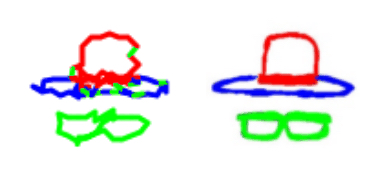} 
\includegraphics[width=0.16\linewidth]{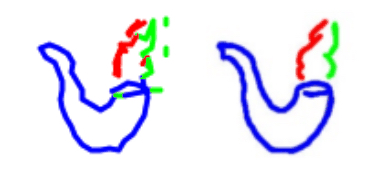}

\includegraphics[width=0.16\linewidth]{colors/13} 
\includegraphics[width=0.16\linewidth]{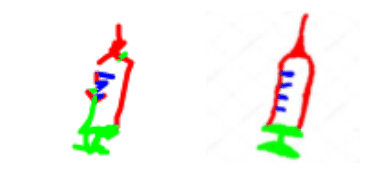} 
\includegraphics[width=0.16\linewidth]{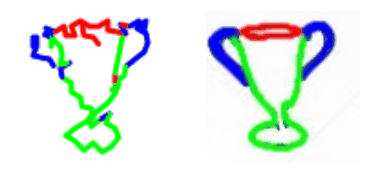}
\includegraphics[width=0.16\linewidth]{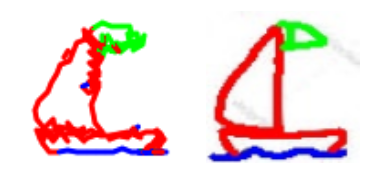}
\includegraphics[width=0.16\linewidth]{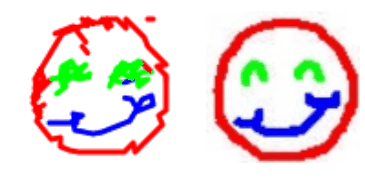}
\includegraphics[width=0.16\linewidth]{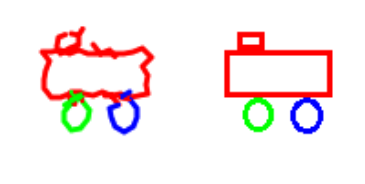} 

\includegraphics[width=0.16\linewidth]{colors/0} 
\includegraphics[width=0.16\linewidth]{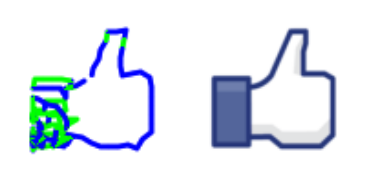} 
\includegraphics[width=0.16\linewidth]{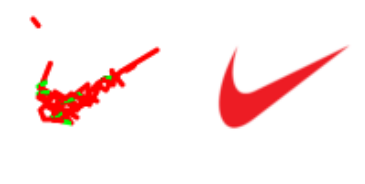} 
\includegraphics[width=0.16\linewidth]{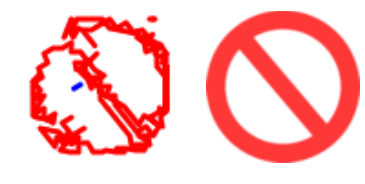}
\includegraphics[width=0.16\linewidth]{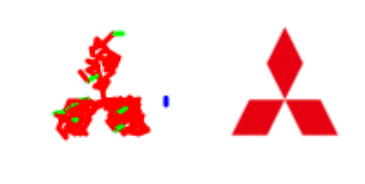} 
\includegraphics[width=0.16\linewidth]{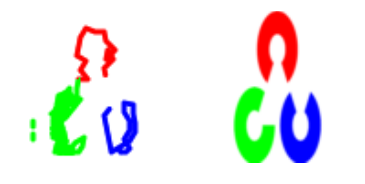}

\includegraphics[width=0.325\linewidth]{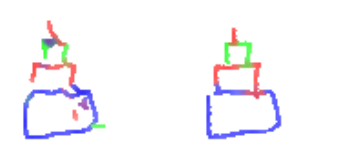}
\includegraphics[width=0.325\linewidth]{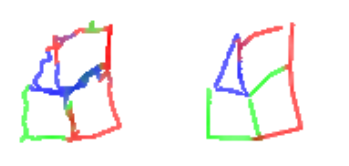} 
\includegraphics[width=0.325\linewidth]{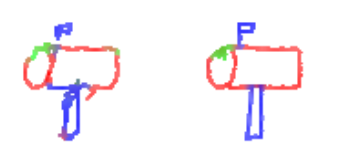}\\

\includegraphics[width=0.325\linewidth]{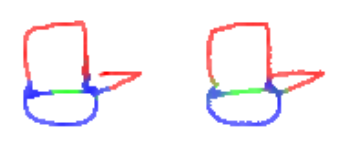}
\includegraphics[width=0.325\linewidth]{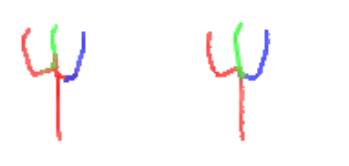}
\includegraphics[width=0.325\linewidth]{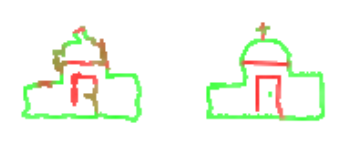}

\end{center}
\caption{Additional sketch drawing examples.}
\label{fig:Grayscale}
\end{figure}

To visualize the effect of the algorithm, the model is unrolled for 100 steps starting from an empty canvas. We chose 100 steps because more steps do not lead to further improvement. Figure~\ref{fig:result} shows the drawing given the reference images from different categories in the test set using different media types. Additional sketch drawing examples are presented (Figure~\ref{fig:Grayscale}) and the algorithm was tested on reference images not in the QuickDraw dataset, where we found that, although it was trained on QuickDraw, the agent has the ability to draw quite diverse doodles. For a reference image, the reward from each step is summed up and the accumulated reward is a quantitative measure of the performance of the algorithm. The maximum reward is achieved when the agent perfectly reproduces the reference image. In the test phase, we used 100 house reference images and 100 reference images randomly selected from the test sets belonging to the training classes. 

\begin{table}[ht]
	\small
	\centering
	\begin{tabular}{|c|c|c|c|c|c|c|c|}
		\hline
		& & \begin{tabular}{@{}c@{}}Naive \\ SDQ \end{tabular}
		& \begin{tabular}{@{}c@{}}SDQ $+$ \\ Rare exp\end{tabular}
		& \begin{tabular}{@{}c@{}}Pretrain \\ on \\ random\end{tabular}
		& \begin{tabular}{@{}c@{}}Pretrain \\ on \\ QuickDraw\end{tabular}
		& \begin{tabular}{@{}c@{}}SDQ $+$ \\ Rare exp $+$ \\ weight init\end{tabular}
		& \begin{tabular}{@{}c@{}}Max \\ reward\end{tabular}\\
		\hline
		\multirow{3}{3em}{House Class}&\begin{tabular}{@{}c@{}}Sketch \end{tabular} & 93 & 1,404 & 1,726 & 1,738 & \color{blue}{1,927} & 2,966\\
	
		&\begin{tabular}{@{}c@{}}Color Sketch \end{tabular} & -13 & 1,651 & 1,765 & 1,747 & \color{blue}{1},808 & 3,484\\
	
		&\begin{tabular}{@{}c@{}}Water Color \end{tabular} &-162 &407 &596 &620 &\color{blue}{670} &1,492 \\
		\hline
		\multirow{3}{3em}{Training Classes}&\begin{tabular}{@{}c@{}}Sketch \end{tabular} & 67 & 1024 & 1,539 & 1,521 & \color{blue}{1,805} & 2,645\\
		
		&\begin{tabular}{@{}c@{}}Color Sketch \end{tabular} & -15 & 1,464 & 1,669 & 1,683 & \color{blue}{1,731} & 3,533\\
		
		&\begin{tabular}{@{}c@{}}Water Color \end{tabular} &$-182$ &363 &446 &473 &\color{blue}{509} &1527 \\
		\hline
	\end{tabular}
	\bigskip
	\caption{Average accumulated rewards for the models tested.}
	\label{tbl:rst}
\end{table}

Table~\ref{tbl:rst} presents the average accumulated rewards and the average maximum rewards across reference images. In the table, the `Naive SDQ' model is the Doodle-SDQ model trained from scratch following a $\epsilon$-greedy strategy with $\epsilon$ annealed linearly from 1.0 to 0.1 over the first fifty thousand frames and fixed at 0.1 thereafter. The `SDQ $+$ Rare exp' is the Doodle-SDQ model trained from scratch with rare exploration. The `Pretrain on random' model is the model with supervised pretraining on the synthesized random stroke sequence data (Figure~\ref{fig:DP}). The `Pretrain on QuickDraw' model is the model with supervised pretraining on the QuickDraw sequence data. The `SDQ $+$ Rare exp $+$ weight init' model is the Doodle-SDQ model with rare exploration and weight initialization from the `Pretrain on random' model. Based on the average accumulated reward, Doodle-SDQ with weight initialization is significantly better than all the other methods. Furthermore, pretraining on the QuickDraw sequence data directly does not lead to superior performance over the RL method. This indicates the advantage of using DRL in the drawing task.

\section{Discussion}

We now list several key factors that contribute to the success of our Doodle-SDQ algorithm and compare it to the DDQN $+$ PER model of \citet{schaul2016prioritized} (Table~\ref{table:summary}).

Since Naive SDQ cannot be directly used for the drawing task, we first pretrain the network to initialize the weights. Referring to Table~\ref{tbl:rst}, pretraining with stroke demonstration via supervised learning leads to an improvement in performance (Columns~4 and 7). Based on our observations, the 4-frame history used in \cite{schaul2016prioritized} introduces a movement momentum that compels the agent to move in a straight line and rarely turn. Therefore, history information is excluded in our current model. In \cite{schaul2016prioritized}, the probability for the exploration of random action decays from 0.9 to 0.1 with increasing epochs. Since we pretrained the network, the agent does not need to explore the environment with a large rate~\cite{cruz2017pre}. Thus, we initially set the exploration rate to 0.1. However, Doodle-SDQ cannot outperform the pretrained model until we remove exploration.\footnote{A random movement will be generated only when the agent gets stuck at some position, such as moving back and forth or remaining at the same spot.} The countereffect of the exploration may be caused by the large action space. The small patch in the two streams structure (Figure~\ref{fig:framework}) makes the agent attend to the region where the pen is located. More specifically, when the lifted pen is within one step action distance to the target drawing, the local stream is able to move the pen to a correct position and start drawing. Without this stream, the RL training cannot be successful even after removing the exploration or pretraining the network. The average accumulated rewards for the global stream only network varies around 100 depending on the media types.

\begin{table}[ht]
\small
\begin{center}
\begin{tabular}{|l|c|c|}
\hline
 & Doodle-SDQ & DDQN $+$ PER \cite{schaul2016prioritized}\\
\hline
		History & No & Yes \\  
		Exploration & Rare & Yes \\  
		Pretrain & Yes & No \\
		Input stream & 2 & 1 \\
\hline
\end{tabular}
\end{center}
\caption{Differences between the proposed method and \cite{schaul2016prioritized}.}
\label{table:summary}
\end{table}

Despite the success of our SDQ model in simple sketch drawing, there are several limitations to be addressed in the future work. On the one hand, the motivation of this paper is to design an algorithm to enable machines to doodle like humans, rather than competing with GAN~\cite{goodfellow2014generative} to generate complex image types, at least not at the current stage. However, it has been demonstrated that an adversarial framework~\cite{ganin2018synthesizing} interprets and generates images in the space of visual programs. Therefore, it will be a promising direction to mimic human drawing by combining adversarial training technique and reinforcement learning. On the other hand, although the SDQ model works in a relatively large action space due to rare exploration, the average accumulated rewards introduced by the component of reinforcement learning still suffers from the increase of the dimension of action space by allowing colorful drawing as shown by a comparison between sketch and color sketch (Column 6 and 7 in Table~\ref{tbl:rst}).
Since our future work will incorporate more action variables (e.g., the pen's pressure and additional colors) and explore doodling on large canvases, the actions might be embed in a continuous space~\cite{dulac2015deep}.

\section{Conclusion}

In this paper we addressed the challenging problem of emulating human doodling. To solve this problem, we proposed a deep-reinforcement-learning-based method, Doodle-SDQ. Due to the large action space, Naive SDQ fails to draw appropriately. Thus, we designed a hybrid approach that combines supervised imitation learning and reinforcement learning. We train the agent in a supervised manner by providing demonstration strokes with ground truth actions. We then further trained the pre-trained agent with Q-learning using a reward based on the similarity between the current drawing and the reference image. Drawing step-by-step, our model reproduces reference images by comparing the similarity between the current drawing and the reference image. Our experimental results demonstrate that our model is robust and generalizes to classes not presented during training, and that it can be easily extended to other media types, such as watercolor.
\newpage
\bibliography{bmvc18}

\begin{thebibliography}{31}
\providecommand{\natexlab}[1]{#1}
\providecommand{\url}[1]{\texttt{#1}}
\expandafter\ifx\csname urlstyle\endcsname\relax
  \providecommand{\doi}[1]{doi: #1}\else
  \providecommand{\doi}{doi: \begingroup \urlstyle{rm}\Url}\fi

\bibitem[Abadi et~al.(2016)Abadi, Barham, Chen, Chen, Davis, Dean, Devin,
  Ghemawat, Irving, Isard, et~al.]{abadi2016tensorflow}
Mart{\'\i}n Abadi, Paul Barham, Jianmin Chen, Zhifeng Chen, Andy Davis, Jeffrey
  Dean, Matthieu Devin, Sanjay Ghemawat, Geoffrey Irving, Michael Isard, et~al.
\newblock Tensorflow: A system for large-scale machine learning.
\newblock In \emph{OSDI}, volume~16, pages 265--283, 2016.

\bibitem[Brown(2014)]{brown2014doodle}
Sunni Brown.
\newblock \emph{The Doodle revolution: Unlock the power to think differently}.
\newblock Penguin, 2014.

\bibitem[Cruz~Jr et~al.(2017)Cruz~Jr, Du, and Taylor]{cruz2017pre}
Gabriel~V Cruz~Jr, Yunshu Du, and Matthew~E Taylor.
\newblock Pre-training neural networks with human demonstrations for deep
  reinforcement learning.
\newblock \emph{arXiv preprint arXiv:1709.04083}, 2017.

\bibitem[Denton et~al.(2015)Denton, Chintala, Fergus, et~al.]{denton2015deep}
Emily~L Denton, Soumith Chintala, Rob Fergus, et~al.
\newblock Deep generative image models using a laplacian pyramid of adversarial
  networks.
\newblock In \emph{Advances in neural information processing systems}, pages
  1486--1494, 2015.

\bibitem[Dulac-Arnold et~al.(2015)Dulac-Arnold, Evans, van Hasselt, Sunehag,
  Lillicrap, Hunt, Mann, Weber, Degris, and Coppin]{dulac2015deep}
Gabriel Dulac-Arnold, Richard Evans, Hado van Hasselt, Peter Sunehag, Timothy
  Lillicrap, Jonathan Hunt, Timothy Mann, Theophane Weber, Thomas Degris, and
  Ben Coppin.
\newblock Deep reinforcement learning in large discrete action spaces.
\newblock \emph{arXiv preprint arXiv:1512.07679}, 2015.

\bibitem[Elgammal et~al.(2017)Elgammal, Liu, Elhoseiny, and
  Mazzone]{elgammal2017can}
Ahmed Elgammal, Bingchen Liu, Mohamed Elhoseiny, and Marian Mazzone.
\newblock Can: Creative adversarial networks, generating" art" by learning
  about styles and deviating from style norms.
\newblock \emph{arXiv preprint arXiv:1706.07068}, 2017.

\bibitem[Ganin et~al.(2018)Ganin, Kulkarni, Babuschkin, Eslami, and
  Vinyals]{ganin2018synthesizing}
Yaroslav Ganin, Tejas Kulkarni, Igor Babuschkin, SM~Eslami, and Oriol Vinyals.
\newblock Synthesizing programs for images using reinforced adversarial
  learning.
\newblock \emph{arXiv preprint arXiv:1804.01118}, 2018.

\bibitem[Gatys et~al.(2016)Gatys, Ecker, and Bethge]{gatys2016image}
Leon~A Gatys, Alexander~S Ecker, and Matthias Bethge.
\newblock Image style transfer using convolutional neural networks.
\newblock In \emph{Proceedings of the IEEE Conference on Computer Vision and
  Pattern Recognition}, pages 2414--2423, 2016.

\bibitem[Goodfellow et~al.(2014)Goodfellow, Pouget-Abadie, Mirza, Xu,
  Warde-Farley, Ozair, Courville, and Bengio]{goodfellow2014generative}
Ian Goodfellow, Jean Pouget-Abadie, Mehdi Mirza, Bing Xu, David Warde-Farley,
  Sherjil Ozair, Aaron Courville, and Yoshua Bengio.
\newblock Generative adversarial nets.
\newblock In \emph{Advances in neural information processing systems}, pages
  2672--2680, 2014.

\bibitem[Graves(2013)]{graves2013generating}
Alex Graves.
\newblock Generating sequences with recurrent neural networks.
\newblock \emph{arXiv preprint arXiv:1308.0850}, 2013.

\bibitem[Gregor et~al.(2015)Gregor, Danihelka, Graves, Rezende, and
  Wierstra]{gregor2015draw}
Karol Gregor, Ivo Danihelka, Alex Graves, Danilo~Jimenez Rezende, and Daan
  Wierstra.
\newblock Draw: A recurrent neural network for image generation.
\newblock \emph{arXiv preprint arXiv:1502.04623}, 2015.

\bibitem[Ha and Eck(2017)]{ha2017neural}
David Ha and Douglas Eck.
\newblock A neural representation of sketch drawings.
\newblock \emph{arXiv preprint arXiv:1704.03477}, 2017.

\bibitem[Hasselt et~al.(2016)Hasselt, Guez, and Silver]{van2016deep}
Hado~van Hasselt, Arthur Guez, and David Silver.
\newblock Deep reinforcement learning with double q-learning.
\newblock In \emph{AAAI Conference on Artificial Intelligence}, pages
  2094--2100. AAAI Press, 2016.

\bibitem[Hester et~al.(2018)Hester, Vecerik, Pietquin, Lanctot, Schaul, Piot,
  Sendonaris, Dulac-Arnold, Osband, Agapiou, et~al.]{hester2018deep}
Todd Hester, Matej Vecerik, Olivier Pietquin, Marc Lanctot, Tom Schaul, Bilal
  Piot, Andrew Sendonaris, Gabriel Dulac-Arnold, Ian Osband, John Agapiou,
  et~al.
\newblock Deep q-learning from demonstrations.
\newblock \emph{Association for the Advancement of Artificial Intelligence
  (AAAI)}, 2018.

\bibitem[Hussein et~al.(2017)Hussein, Gaber, Elyan, and
  Jayne]{hussein2017imitation}
Ahmed Hussein, Mohamed~Medhat Gaber, Eyad Elyan, and Chrisina Jayne.
\newblock Imitation learning: A survey of learning methods.
\newblock \emph{ACM Computing Surveys (CSUR)}, 50\penalty0 (2):\penalty0 21,
  2017.

\bibitem[J et~al.(2016)J, H, T, J, and N.]{Jongejan2016quick}
Jongejan J, Rowley H, Kawashima T, Kim J, and Fox-Gieg N.
\newblock The quick, draw! - ai experiment.
\newblock \emph{https://quickdraw.withgoogle.com}, 2016.

\bibitem[Kingma and Ba(2014)]{kingma2014adam}
Diederik Kingma and Jimmy Ba.
\newblock Adam: A method for stochastic optimization.
\newblock \emph{arXiv preprint arXiv:1412.6980}, 2014.

\bibitem[Kingma and Welling(2013)]{kingma2013auto}
Diederik~P Kingma and Max Welling.
\newblock Auto-encoding variational bayes.
\newblock \emph{arXiv preprint arXiv:1312.6114}, 2013.

\bibitem[Levine et~al.(2016)Levine, Finn, Darrell, and Abbeel]{levine2016end}
Sergey Levine, Chelsea Finn, Trevor Darrell, and Pieter Abbeel.
\newblock End-to-end training of deep visuomotor policies.
\newblock \emph{The Journal of Machine Learning Research}, 17\penalty0
  (1):\penalty0 1334--1373, 2016.

\bibitem[Mnih et~al.(2015)Mnih, Kavukcuoglu, Silver, Rusu, Veness, Bellemare,
  Graves, Riedmiller, Fidjeland, Ostrovski, Petersen, Beattie, Sadik,
  Antonoglou, King, Kumaran, Wiestra, Legg, and Hassabis]{mnih2015human}
Volodymyr Mnih, Koray Kavukcuoglu, David Silver, Andrei~A. Rusu, Joel Veness,
  Marc~G. Bellemare, Alex Graves, Martin Riedmiller, Andreas~K. Fidjeland,
  Georg Ostrovski, Stig Petersen, Charles Beattie, Amir Sadik, Ioannis
  Antonoglou, Helen King, Dharshan Kumaran, Dan Wiestra, Shane Legg, and Demis
  Hassabis.
\newblock Human-level control through deep reinforcement learning.
\newblock \emph{Nature}, 518\penalty0 (7540):\penalty0 529--33, 2015.

\bibitem[Peng et~al.(2016)Peng, Berseth, and Van~de Panne]{peng2016terrain}
Xue~Bin Peng, Glen Berseth, and Michiel Van~de Panne.
\newblock Terrain-adaptive locomotion skills using deep reinforcement learning.
\newblock \emph{ACM Transactions on Graphics (TOG)}, 35\penalty0 (4):\penalty0
  81, 2016.

\bibitem[Ross et~al.(2011)Ross, Gordon, and Bagnell]{ross2011reduction}
St{\'e}phane Ross, Geoffrey Gordon, and Drew Bagnell.
\newblock A reduction of imitation learning and structured prediction to
  no-regret online learning.
\newblock In \emph{Proceedings of the fourteenth international conference on
  artificial intelligence and statistics}, pages 627--635, 2011.

\bibitem[Schaul et~al.(2016)Schaul, Quan, Antonoglou, and
  Silver]{schaul2016prioritized}
T.~Schaul, J.~Quan, I.~Antonoglou, and D.~Silver.
\newblock Prioritized experience replay.
\newblock In \emph{International Conference on Learning Representations
  (ICLR)}, 2016.

\bibitem[Silver et~al.(2016)Silver, Huang, Maddison, Guez, Sifre, van~den
  Driessche, Schrittwieser, Antonoglou, Panneershelvam, Lanctot, Dieleman,
  Grewe, Nham, Kalchbrenner, Sutskever, Lillicrap, Leach, Kavukcuoglu, Graepel,
  and Hassabis]{silver2016mastering}
David Silver, Aja Huang, Chris~J. Maddison, Arthur Guez, Laurent Sifre, George
  van~den Driessche, Julian Schrittwieser, Ioannis Antonoglou, Veda
  Panneershelvam, Marc Lanctot, Sander Dieleman, Dominik Grewe, John Nham, Nal
  Kalchbrenner, Ilya Sutskever, Timothy Lillicrap, Madeleine Leach, Koray
  Kavukcuoglu, Thore Graepel, and Demis Hassabis.
\newblock Mastering the game of go with deep neural networks and tree search.
\newblock \emph{Nature}, 529\penalty0 (7587):\penalty0 484--489, 2016.

\bibitem[Simhon and Dudek(2004)]{simhon2004sketch}
Saul Simhon and Gregory Dudek.
\newblock Sketch interpretation and refinement using statistical models.
\newblock In \emph{Rendering Techniques}, pages 23--32, 2004.

\bibitem[Subramanian et~al.(2016)Subramanian, Isbell~Jr, and
  Thomaz]{subramanian2016exploration}
Kaushik Subramanian, Charles~L Isbell~Jr, and Andrea~L Thomaz.
\newblock Exploration from demonstration for interactive reinforcement
  learning.
\newblock In \emph{Proceedings of the 2016 International Conference on
  Autonomous Agents \& Multiagent Systems}, pages 447--456. International
  Foundation for Autonomous Agents and Multiagent Systems, 2016.

\bibitem[Sun et~al.(2014)Sun, Qian, and Xu]{sun2014robot}
Yuandong Sun, Huihuan Qian, and Yangsheng Xu.
\newblock Robot learns chinese calligraphy from demonstrations.
\newblock In \emph{Intelligent Robots and Systems (IROS 2014), 2014 IEEE/RSJ
  International Conference on}, pages 4408--4413. IEEE, 2014.

\bibitem[Tresset and Leymarie(2013)]{tresset2013portrait}
Patrick Tresset and Frederic~Fol Leymarie.
\newblock Portrait drawing by paul the robot.
\newblock \emph{Computers \& Graphics}, 37\penalty0 (5):\penalty0 348--363,
  2013.

\bibitem[Vondrick et~al.(2016)Vondrick, Pirsiavash, and
  Torralba]{vondrick2016generating}
Carl Vondrick, Hamed Pirsiavash, and Antonio Torralba.
\newblock Generating videos with scene dynamics.
\newblock In \emph{Advances In Neural Information Processing Systems}, pages
  613--621, 2016.

\bibitem[Xie et~al.(2012)Xie, Hachiya, and Sugiyama]{xie2012artist}
Ning Xie, Hirotaka Hachiya, and Masashi Sugiyama.
\newblock Artist agent: a reinforcement learning approach to automatic stroke
  generation in oriental ink painting.
\newblock In \emph{Proceedings of the 29th International Coference on
  International Conference on Machine Learning}, pages 1059--1066. Omnipress,
  2012.

\bibitem[Zhang et~al.(2017)Zhang, Yin, Zhang, Liu, and
  Bengio]{zhang2017drawing}
Xu-Yao Zhang, Fei Yin, Yan-Ming Zhang, Cheng-Lin Liu, and Yoshua Bengio.
\newblock Drawing and recognizing chinese characters with recurrent neural
  network.
\newblock \emph{IEEE transactions on pattern analysis and machine
  intelligence}, 2017.

\end{thebibliography}

\end{document}